\documentclass[sigchi]{acmart}
\usepackage[ruled,norelsize]{algorithm2e}
\usepackage{epsfig}
\usepackage{graphicx}
\usepackage{amsmath}
\usepackage{amssymb}
\usepackage{amsfonts}
\usepackage{color}
\usepackage{colortbl}
\usepackage{multirow}
\usepackage{subcaption}
\usepackage{scrextend}
\usepackage{tabularx}
\usepackage{booktabs}
\usepackage{bbm}
\usepackage{breqn}
\makeatletter
\newcommand{\removelatexerror}{\let\@latex@error\@gobble}
\makeatother

\begin{document}

\title{Multi-modal Active Learning From Human Data:\\A Deep Reinforcement Learning Approach }

\author{Ognjen (Oggi) Rudovic$^{1}$, Meiru Zhang$^{2}$, Bj{\"o}rn Schuller$^{2}$ and Rosalind W. Picard$^1$}

\affiliation{{\small  $^1$Massachusetts Institute of Technology, USA, $^2$Imperial College London, UK}\\{\tt\small \{orudovic,roz\}@mit.edu, \,\,\{meiru.zhang18, bjoern.schuller\}@imperial.ac.uk}}
\renewcommand{\shortauthors}{Rudovic, et al.}
\renewcommand{\shorttitle}{Multi-modal Active Learning From Human Data}

\begin{abstract}
  Human behavior expression and experience are inherently multimodal, and characterized by vast individual and contextual heterogeneity. To achieve meaningful human-computer and human-robot interactions, multi-modal models of the user's states (e.g., engagement) are therefore needed. Most of the existing works that try to build classifiers for the user's states assume that the data to train the models are fully labeled. Nevertheless, data labeling is costly and tedious, and also prone to subjective interpretations by the human coders. This is even more pronounced when the data are multi-modal (e.g., some users are more expressive with their facial expressions, some with their voice). Thus, building models that can accurately estimate the user's states during an interaction is challenging. To tackle this, we propose a novel multi-modal active learning (AL) approach that uses the notion of deep reinforcement learning (RL) to find an optimal policy for active selection of the user's data, needed to train the target (modality-specific) models. We investigate different strategies for multi-modal data fusion, and show that the proposed model-level fusion coupled with RL outperforms the feature-level and modality-specific models, and the na\"ive AL strategies such as random sampling, and the standard heuristics such as uncertainty sampling. We show the benefits of this approach on the task of engagement estimation from real-world child-robot interactions during an autism therapy. Importantly, we show that the proposed multi-modal AL approach can be used to efficiently personalize the engagement classifiers to the target user using a small amount of actively selected user's data. 
\end{abstract}


\maketitle

\section{Introduction}
Human behavior is inherently multi-modal, and individuals use eye gaze, hand gestures, facial expressions, body posture, and tone of voice along with speech to convey engagement and regulate social interactions~\cite{el2006affective}. A large body of work in human-computer interaction (HCI) and human-robot interaction (HRI) explored the use of various affective and social cues to facilitate and assess the user engagement~\cite{goodrich2008human,tsiourti2019multimodal}. Most of these works can be divided into those that detect the presence of a set of the engagement cues or interaction events~\cite{chang2018ensemble,park2019aaai,gordon_affective_2016}, or use supervised classifiers trained with social, physiological, or task-based interaction features~\cite{bohus2014managing,sanghvi2011automatic,castellano2009detecting}.  
\begin{figure*}[t]
    \centering
 \includegraphics[width=0.75\textwidth]{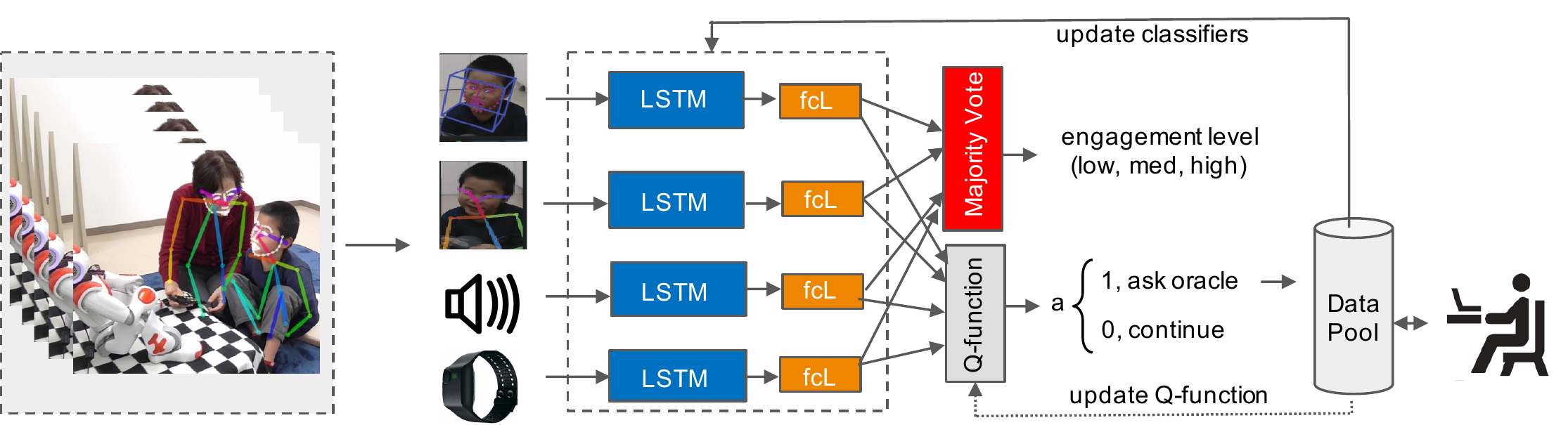}
\caption{Overview of the proposed multi-modal AL approach. The input are the recordings of the child-robot interactions during an autism therapy (we used the camera placed behind the robot). The image frames are first processed using open-source tools (openFace~\cite{openface} and openPose~\cite{openpose}) to obtain the facial and body cues. Likewise, the audio recordings are processed using the openSMILE toolkit~\cite{opensmile}. We also used the data collected by the E4 wristband~\cite{e4} on the child's hand (providing autonomic physiology data such as galvanic skin conductance and body temperature, as well as the accelerometer data). These data are fed as input to the modality-specific engagement classifiers: the LSTM models followed by fully-connected layers (fcL). The classification of the engagement levels (low, medium, high) is performed by applying majority voting to the classifiers' outputs. These are also fed into the Q-function of the RL policy for active data-selection (also modeled using an LSTM cell and fcL for the action selection: to ask or not ask for the label). If the label is requested, the input multi-modal data is stored in a data pool for labelling by the human expert. These data are then used to train the data-selection policy and target classifiers. During inference of the data of a new child, the actively-selected data are used to personalize the engagement classifiers.}
\label{fig:model}
\end{figure*}
A detailed overview of related multi-modal approaches can be found in~\cite{alameda2019multimodal,pantic2003toward}. The majority of these approaches adopt either feature-level (e.g., by simple concatenation of the multi-modal input features) or model-level (e.g., by combining multiple classifiers trained for each data modality) fusion~\cite{baltruvsaitis2019multimodal}. While this can improve the estimation of target outcomes (e.g., the user engagement) when compared to the single-modality classifiers, these methods usually adopt "one-size-fits-all" learning approach where the trained models are applied to new users without the adaptation to the user. Consequently, their performance is usually limited when the users' data are highly heterogeneous (e.g., due to the differences in facial expressions/body gestures as a result of individual engagement styles). 

Recently, several works proposed models for personalized estimation of engagement in HRI. For instance,~\cite{rudovic2018personalized} proposed a multi-modal deep learning for engagement estimation that combines body, face, audio and autonomic physiology data of children with autism during the therapy sessions with a humanoid robot. 
However, most of multi-modal works in HCI and HRI are fully supervised~\cite{baltruvsaitis2019multimodal,kahou2016emonets,jaimes2007multimodal}, i.e., they assume that the data used to train the models are fully labeled. Obtaining these labels is expensive, especially when dealing with a large amount of audio-visual recordings. Therefore, there is a need for methods that can automatically select the most informative instances that need to be labeled in order to train the target estimation models. More importantly, to improve the generalization of these models, we need to select the data of a new user that can be used to personalize the target models. Yet, how to select the most informative instances of the multi-modal data from the target user is an open research problem that has not been investigated much.   

To address this, the approach proposed here uses the notion of AL~\cite{settles2010active}. Central to the AL framework is the query strategy used to decide when to request a label for target data. The most commonly used query strategies include uncertainty sampling, entropy, or query-by-committee~\cite{settles2010active}. Furthermore, more advanced query strategies have been proposed to adapt deep network classifiers based on the uncertainty of the network output (e.g.,~\cite{kading2016active,wang2016cost,konyushkova2017learning}). Yet, the candidate query strategies still must be specified by a human. More importantly, there is not one strategy that works the best for all users. Instead of using the heuristic strategies, recent (deep) AL approaches (e.g., ~\cite{liu2018learning,fang2017learning,woodward2017active,wang2016learning,duan2016rl}) have adopted a data-driven approach that learns a model-free AL off-line policy using RL~\cite{sutton1998}. For instance,~\cite{woodward2017active} proposed a model where an agent makes a decision whether to request a label or make a prediction. The agent receives a reward related to its decision: a positive reward is given for correct predictions, and negative rewards for incorrect predictions or label requests. This can be achieved by the Q-function modeled using the notion of deep RL~\cite{mnih2013playing,fang2017learning}. The main goal of these approaches is to adapt the prediction model to new tasks, using a minimum number of queries. Our work is a generalization of the RL framework for AL~\cite{fang2017learning} to multi-modal data, where instead of dealing with a single agent-environment, the agent deals simultaneously with multiple environments (i.e., data modalities).

Note that RL has previously been applied in the tasks of multi-modal learning. For instance,~\cite{qian2018multimodal} used RL to enhance the machine translation from different data modalities. Likewise,~\cite{jiang2017deep} used RL for image question answering, an inherently multi-modal learning problem. Also, in the context of visual dialogues, RL has been used with multi-modal learning~\cite{zhang2018multimodal}. However, none of these works explored RL for AL from multimodal data. The most related approach to ours is the multi-view AL framework~\cite{muslea2006active}. It uses the standard heuristic AL strategies to select data from multiple views. While different views can be seen as different modalities of the same phenomenon, like facial and body gestures of human behaviour, to our knowledge no previous work has attempted multi-modal AL from the real-world human-interaction data and using RL to learn an optimal data selection policy. Moreover, the model personalization using such approach has not been explored before. 

To tackle the challenges of learning from multi-modal human data (as typically encountered in HCI and HRI), in this work we formulate a novel multi-modal AL approach. We show that this approach can be used to personalize the data-modality-specific classifiers to the target user using a small amount of labeled data of the user, which are automatically selected using the newly proposed multi-modal AL strategy. The main contributions of this work are: (i) We propose a novel approach for multi-modal AL using RL for training a policy for active data-selection. 
(ii) We propose a novel personalization strategy based on the actively-selected multi-modal data of the target user. (iii) We show on a highly challenging dataset of child-robot interactions during an autism therapy that the proposed approach leads to large improvements in estimation of engagement (low, medium, high) from the multi-modal data, when compared to non-personalized models, and heuristic AL strategies. The outline of the proposed approach is depicted in Fig.~\ref{fig:model}. Compared to traditional supervised classifiers for multi-modal data, our approach provides an efficient mechanism for actively selecting the most relevant data for training the target engagement classifier, thus,  minimizing the human data-labelling efforts.

\section{Preliminaries}
\subsection{Problem Statement and Notation}
In our learning setting, we use multi-modal data recordings of child-robot interactions during an autism therapy~\cite{rudovic2018personalized}, as described in Sec.~\ref{sec:exp}. Formally, we denote our dataset as ${\mathcal{D}} = \{ D^{(1)},..,D^{(m)},..,D^{(M)}\}$, where $m=1,..,M$ denotes the data modality (e.g., face, body, etc.). This dataset comprises video recordings of target interactions of $C$ children (later split into training and test child-independent partitions). We assume a single recording per child, which may vary in duration. Each recording is represented with a set of multi-modal features $\{X,Y\}$, where $X=[x_1,..,x_N]$, with $x_i\in \mathcal{R}^{T\times M \times D_M}$ containing the collection of the multi-modal features extracted every 60 ms from a sliding window of 1 second duration (30 image frames), resulting in $T=10$ temporally correlated multi-modal feature representations (the dimension of modality $m=1,..,M$ is $D_m$). The features $x_i$ of each 1 second interval are associated with the target engagement label $y_i=\{0,1,2\}$ (see Sec.~\ref{sec:exp} for details). Note that $N$ can vary per child, i.e., per recording. Given these data, we address it as a multi-class multi-modal sequence classification problem, where our goal is two-fold: (i) to predict the target label given the input features extracted from the sliding window within the recording, and (ii) to actively select the data of each child so that our engagement estimation model can be personalized to the target child. 
\subsection{The Base Classification Model}
As the base model for the engagement classifiers ($\varphi$) and also to implement the Q-function in the RL component of our approach, we use the Long Short-Term Memory (LSTM)~\cite{hochreiter1997long} model, which enables long-range learning of time-feature dependencies. This has shown great success in tasks such as action recognition~\cite{donahue2015long,turaga2008machine} and speech analysis~\cite{graves2014towards,eyben2013real}, among others. Each LSTM cell has hidden states augmented with nonlinear mechanisms that allow the network state to propagate without modification, be updated, or be reset, using simple learned gating functions. More formally, a basic LSTM cell can be described with the following equations:
\begin{equation}
\begin{array}{l}
{{\hat g}^f},{{\hat g}^i},{{\hat g}^x},{{\hat c}_t} = {W_x} \cdot {x_t} + {W_h} \cdot {h_{t - 1}} + b\\
{g^f} = \sigma ({{\hat g}^f}),{g^i} = \sigma ({{\hat g}^i}),{g^x} = \sigma ({{\hat g}^x})\\
{c_t} = {g^f} \odot {c_{t - 1}} + {g^i} \odot \tanh ({{\hat c}_t}),\,\,\,\,{h_t} = {g^x} \odot \tanh ({c_t}),
\end{array}
\end{equation}
where ${{\hat g}^f},{{\hat g}^i},{{\hat g}^x}$ are the forget gates, input gates, and output gates respectively, ${{\hat c}_t}$ is the candidate cell state, and $c_t$ is a new LSTM cell state. $W_x$ and $W_h$ are the weights mapping from the observation ($x_t$) and hidden state ($h_{t-1}$), respectively, to the gates and candidate cell state, and $b$ is the bias vector. $\odot$ represents element-wise multiplication; $\sigma(\cdot)$ and $\text{tanh}(\cdot)$ are the sigmoid and hyperbolic tangent functions respectively~\cite{woodward2017active}. To model the window of $T=10$ time steps in $({\bf x}, y)$ data pairs, we feed the feature vectors from ${\bf x}\in [x_1,,..,x_l,..,x_T]$ to the temporally unrolled LSTM cell. Then, their output-state values $h_t$ are passed through fully connected layers fcL$_t$, $t=1,..,T$ (we use a rectified linear unit -- ReLU), and average across the time steps. Finally, a sigmoid layer followed by the softmax is applied to this output to obtain the sequence label $y^*$. 
\subsection{RL for Data-selection Policy Learning}
\label{rlbase}
RL~\cite{sutton1998} is a framework that can be used to learn an optimal policy $\pi$ for actively selecting data samples during model training and adaptation. This  is  a  form  of  Markov  Decision  Process (MDP), which allows the learning of a policy that can dynamically select instances that are most informative~\cite{fang2017learning}. More specifically, given an input feature vector (${\bf x_i}$), we first compute a state-vector ($s_i$) that is then passed to the trained policy (Q-function), which outputs an action ($a_i$). During training of the Q-function, the goal is to maximize the action-value function $Q^*(s_i, a_i)$. This is at the heart of RL, and it specifies the expected sum of discounted future rewards for taking action $a_i$ in state $s_i$ and acting optimally from then on as:
\begin{equation}
{a_i} = {\pi ^*}({s_i}) = \arg {\max _{{a_i}}}\,\, Q^*(s_i, a_i);
\end{equation}
The optimal $Q$ function is given by the Bellman equation:
\begin{equation}
Q^*(s_i, a_i) = \mathbb{E}_{s_{i+1}}[r_i + \gamma \max_{a_{i+1}} Q^*(s_{i+1}, a_{i+1}) | s_i, a_i],
\label{eq:bellman}
\end{equation}
where $\mathbb{E}_{s_{i+1}}$ indicates an expected value over the distribution of possible next states $s_{i+1}$, $r_i$ is the reward at the current state $s_i$ (derived from the input features ${\bf x_i}$), and $\gamma$ is a discount factor, which incentivizes the model to seek reward in fewer time steps. The tuples $(a_i,s_i, r_i,s_{i+1})$ represent a MDP, and they are used to train the Q-function in Eq. (\ref{eq:bellman}).
\section{Methodology}
We propose a novel approach for multi-modal AL that provides an optimal policy for the active data-selection. These data are used consequently to re-train the classification models for estimation of the target output (in our case, the engagement level) from fixed-sized video segments (1 second long). The proposed approach consists of two sequential processes: (i) the training of the classifiers for the target output, and (ii) the learning of the Q-function of the RL model for active data selection. These two are performed in a loop, where first the target classifiers are trained using the data of each modality ($m=1,..,M$), thus, $M$ classifiers are trained in parallel. Then, based on their outputs, we perform the model-level fusion to obtain the target label for the input. This label is used, along with the input features and the true label, to train the Q-function for active data selection. During the models' training, this process is repeated by re-training the classifiers and the Q-function until both models have converged (or for a fixed number of iterations, i.e., training episodes). During inference of new data (in our case, the recording of the interactions between the new child and the robot), the learned group policy (i.e., the Q-function) is first used to select the data samples that the model is uncertain about. Consequently, these are then used to personalize the target classifiers by additionally training each classifier using the actively selected data of the target child. The personalized classifiers are then used to obtain the engagement estimates on the remaining data of the target child. \vspace*{-\baselineskip}

\subsection{Engagement Classifiers}
\label{deepf}
For each data modality ($m=1,..,M$) we train a separate LSTM model, resulting in the following ensemble of the models: $\varphi = \{ \varphi^{(1)},\varphi^{(2)},..,\varphi^{(M)}\}$. Each model is trained using the actively-selected samples from the target modality, where the number of possible samples is defined using a pre-defined budget ($\mathcal{B}$) for active learning. The  number of hidden states in each LSTM was set to 64 (found on a separate validation set), and the size of the fcL was then set to $64\times 3$, where the network output was 1-hot encoded (three engagement levels). Specifically, in the dataset used, we have $M=4$ data modalities, comprising of the FACE ($m=1$), BODY ($m=2$), autonomic physiology (A-PHYS) ($m=3$) and AUDIO ($m=4$) modality. Thus, for the target task, we trained four LSTM models. From each data modality, we used the feature representations extracted using the open-source codes for face and body processing from image data (openFace~\cite{openface} and openPose~\cite{openpose}), as done in~\cite{rudovic2018personalized}. These provide the locations of characteristic facial points, facial action units activations (0/1) and their intensities (0-5), and locations of 18 body joints and their confidence. As features of A-PHYS, we used the data collected with the E4~\cite{e4} wristband on a child's hand, providing the galvanic skin response, heart-rate and body temperature, along with the 3D acelerometer data encoding the child's movements. From the audio recordings, we used 24 low-level descriptors provided by the openSMILE~\cite{opensmile}, an open-source toolkit for audio-feature extraction. The feature dimension per modality was: 257D (FACE), 70D (BODY), 27D (A-PHYS), and 24D (Audio), thus, 378 features in total. For more details about the feature extraction, see~\cite{rudovic2018personalized}. 

To obtain the engagement estimate by the proposed multi-modal approach, we perform the model-level fusion by combining the target predictions of the modality-specific LSTMs:
\begin{equation}
{\widehat y_i}\leftarrow\text{majorityVote}(\varphi^{(1)}({\bf x_i^{(1)}}),\varphi^{(2)}({\bf x_i^{(2)}}),\varphi^{(3)}({\bf x_i^{(3)}}),\varphi^{(4)}({\bf x_i^{(4)}})),
\end{equation}
where in the case of ties, we select the most confident estimate, based on the soft-max outputs of each classifier. We also tried other model-fusion schemes, such as confidence-based weighting, and majority vote on the three most confident estimates, but the basic majority vote performed the best, and was used in the experiments reported.
\subsection{Group-policy Learning for Active Data-selection}
\label{gpolicy}
We first use the training data (i.e., recordings of the children in the training set) to learn the (initial) group-policy $\pi_g$ for making the decision whether to query or not the label for the input features (${\bf x}$). For this, we implement the Q-function using the LSTM model, which receives in its input the states ($s$) and outputs actions ($a$), as described in Sec.~\ref{rlbase}. 
\begin{figure}[!t]
 \removelatexerror
  \begin{algorithm}[H]
   \caption{Multi-modal Q-learning ({\bf MMQL})}
   {\bf Input}: Dataset $D=\{D^{(1)},D^{(2)},\dots,D^{(M)}\}$, models $\left\{ {\varphi _0^{} = \{ \varphi _0^{(1)},\varphi _0^{(2)},..,\varphi _0^{(M)}\} ,\,\,Q_0^\pi } \right\} \leftarrow$ {\it rand}, budget $\mathcal{B}$ \\
   {\bf Output}: Optimized models $\left\{ {\varphi _*^{},\,\,Q_*^\pi } \right\}$\\
   \hrulefill\\
   \For(){$e := 1$ to $N$}
   {
      $D_0^\ell  \leftarrow \emptyset$, shuffle $D$ \;
      $\varphi _e \leftarrow \varphi _{e-1}$, $Q_e^\pi\leftarrow Q_{e-1}^\pi$\;
      \For(){$i := 1$ to $|D|$}
      {${\widehat y_i}\leftarrow\text{majorityVote}(\varphi _e({\bf x_i}))$\;
      $s_i\leftarrow {\bf x_i}$ \# construct a new state\;
      ${a_i} = \mathop {\arg \max }\limits_a Q_e^\pi ({s_i},a)$ \# make a decision\;
      \If(){$a_i==1$}
          {$D_e^\ell\leftarrow D_e^\ell \cup ({\bf x_i},y_i)$ \# ask for label\;
          }
      $r_i\leftarrow R(a_i,{\widehat y_i}, y_i)$ \# compute reward\;
      \If(){$|D_e^\ell|==\mathcal{B}$}
            {store $(s_i, a_i,r_i,{\text end})\rightarrow \mathcal{M}$\;
            break\;
            }
      $s_{i+1}\leftarrow {\bf x_{i+1}}$ \# construct a new state\;
      store $(s_i, a_i, r_i, s_{i+1})\rightarrow \mathcal{M}$\;
      update $Q_{e}^\pi$ using a batch from $\mathcal{M}$\;
      }
      update models $\varphi _e$ using $D_e^\ell$\;
   }
   {\bf Return}: $\varphi _*\leftarrow\varphi _e$, $Q_*^\pi\leftarrow Q_e^\pi$
   \label{mmql}
  \end{algorithm}
\end{figure}

\noindent{\bf States and Actions.}
We approximate the $Q$-function using the LSTM model with 32 hidden units. This is followed by a ReLU layer with the weight and bias parameters $\{W_l,b_l\}$, and softmax in the output, i.e., actions. The actions in our model are binary (ask/ do not ask for label), and are 1-hot encoded using a 2D output vector $a_i$ for input states $s_i$. For instance, if a label is requested, ${a_i}=[1,0]$; otherwise, ${a_i}=[0,1]$ and no data label is provided by an oracle (e.g. the human expert). For training the Q-function, the design of the input states is critical. Typically, the raw input features (in our case, ${\bf x_i}$, are considered to be the states $(s_i)$ of the model~\cite{woodward2017active}. While this is feasible, in the case of multiple modalites, this can lead to a large state-space, which in turn can easily lead to the overfitting of the LSTM used to implement the Q-function. Here, we propose a different approach. Instead of using the input content (${\bf x_i}$) directly, we use the output of the engagement classifiers to form the states of the Q-function. Namely, the sigmoid layer of each target classifier outputs the probabilities for each class label in $y$, i.e., $\varphi^{(m)}\rightarrow (p_1^{(m)},p_2^{(m)},p_3^{(m)})$. To obtain the "content-free"\footnote{Note that this can also be termed as "meta-weights", as in the previously proposed meta-learning AL frameworks, e.g.,~\cite{konyushkova2017learning}.} states for our RL model, for the target input ${\bf x_i}$, we concatenate these estimated probabilities from each classifier to form the state vector $s_i$. Furthermore, we augment this state vector by also adding the overall confidence of each classifier. This is computed as ${C^{(m)}} = 1 - \sum\nolimits_{i = 1}^{|y|} {p_i^{(m)}\log (p_i^{(m)})}$, where the sum on the right-hand side of the equation is the entropy of the classifier (we bound it to $[0,1]$). Finally, in the case of $M=4$, this results in 16D state-vectors $s_i$. As we show in our experiments, such the state representation leads to overall better results and learning of the Q-function than when the raw high-dimensional input features are used to represent the states (in our case, $dim({\bf x})=378$).    

\noindent{\bf Reward.} Another important aspect of RL is the design of the reward function. Given the input multi-modal feature vectors ${\bf x_i}$, the active learner chooses an action $a_i$ of either requesting the true label $y_i$ or not. If the label is requested, the model receives a negative reward to reflect that obtaining labels is costly. On the other hand, if no label is requested, the model receives positive reward if the estimation is correct; otherwise, it receives negative reward. This is encoded by the following RL reward function:
\begin{equation}
r_i(a_i,{\widehat y_i},y_i)=
\begin{cases}
r_{req}=-0.05, \text{if   } a_i=1\\
r_{cor}=+1, \,\,\,\,\,\,\,\,\text{if   } a_i=0 \wedge {\widehat y_i} = y_i\\
r_{inc}=-1, \,\,\,\,\,\,\,\,\text{if   } a_i=0 \wedge {\widehat y_i} \neq y_i\\
\end{cases},
\end{equation}
where $y_i$ is the target label obtained by the majority vote from the modality-specific engagement classifiers (Sec.~\ref{deepf}). This reward is critical for training the $Q$-function that we use to learn the data-selection policy. Note that this type of reward has previously been proposed in~\cite{woodward2017active}, however, it has not been used in the context of multi-modal AL.

\begin{figure}[!t]
 \removelatexerror
  \begin{algorithm}[H]
   \caption{Personalized Engagement Estimation}
   {\bf Input}: New data $D=\{D^{(1)},D^{(2)},\dots,D^{(K)}\}$, models $\left\{ {\varphi _*^{} = \{ \varphi _*^{(1)},\varphi _*^{(2)},..,\varphi _*^{(M)}\} ,\,\,Q_*^\pi } \right\}$, budget $\mathcal{B}$ \\
   {\bf Output}: labeled data in $D$, adapted models $\varphi _*$\\
   \hrulefill\\
   $D^\ell\leftarrow \emptyset$, shuffle $D$\\
   \For(){$i := 1$ to $|D|$}
   {  ${\widehat y_i}\leftarrow\text{majorityVote}(\varphi _*({\bf x_i}))$\;
      $s_i\leftarrow {\bf x_i}$ \# construct a new state\;
      ${a_i} = \mathop {\arg \max }\limits_a Q_*^\pi ({s_i},a)$ \# make a decision\;
      \If(){$a_i==1$}
          {$D^\ell\leftarrow D^\ell \cup ({\bf x_i},y_i)$ \# ask for label\;
          $D\leftarrow D \setminus {\bf x_i}$\;
          }
      \If(){$|D^\ell|==\mathcal{B}$}
      {break\;}
    }
  update models $\varphi_*$ using $D^\ell$\;
  make estimates ${\widehat Y}\leftarrow\text{majorityVote}(\varphi _*({\bf X}\in D))$\;
  {\bf Return}: ${\widehat Y}\in D, Y\in D^\ell, \varphi _*$
  \label{pmmql}
  \end{algorithm}
\end{figure}
\noindent{\bf Optimization.} Given the space-action pairs, along with the rewards, the parameters of the $Q$-function are optimized by minimizing the Bellman loss on the training data:
\begin{equation}
L_B^{(i)}(\Theta)=[Q_{\Theta}(s_i, a_i) - (r_i + \gamma \max_{a_{i+1}} Q_\Theta(s_{i+1}, a_{i+1}))]^2,
 \label{eq:bellmanloss}
\end{equation}
which encourages the model to improve its estimate of the expected reward at each training iteration (we set $\gamma=0.9$). This is performed sequentially over $i=1,\dots, N$ multi-modal data samples from the training children, and over a number of training episodes. The loss minimizaton is performed using Adam optimizer with the learning rate 0.001. The training of this new AL approach, named the Multi-modal Q-learning (MMQL), is summarized in Alg.~\ref{mmql}. 
\begin{table*}
\small
\caption{The distribution of the engagement levels for each test child. The data samples are obtained by applying a 1-second sliding window, with a shift of 20 ms, to the original recordings of the target children.}
\begin{tabular}{l|c|c|c|c|c|c|c|c|c|c|c|c|c|c|}
\hline
\hline
\textbf{Test child ID} &1 &2 &3 &4 &5 &6 &7 &8 &9 &10 &11 &12 &13 &14 \\ \hline
\textbf{\# of samples in class 0} &697 &0 &72 &31 &541 &2281 &3903 &478 &125 &0  &188 &20 &0 &1134\\
\textbf{\# of samples in class 1} &683 &0 &107 &72 &241 &1026 &1004 &772 &318 &0 &101 &204 &107 &44\\
\textbf{\# of samples in class 2} &0 &2539 &2095 &1452 &0 &0 &898 &87 &549 &1689 &1524 &5686 &2107 &0\\
\hline
\hline
\end{tabular}
\label{tab_views_3}
\end{table*}
\begin{table*}
\small
\caption{ACC [\%] of the models before ({\it left}) and after ({\it right}) adaptation to test children using different budgets (5, 10, 20, 50 and 100) for active data-selection (averaged across the children and budgets).}
\begin{tabular}{l|c|c|c|c|c|c|}
\hline
\hline
{\textbf{Model}} &\textbf{FACE} &\textbf{BODY} &\textbf{A-PHYS} &\textbf{AUDIO}  &\textbf{FEATURE-F} &\textbf{MODEL-F} \\ \hline
\textbf{MMQL (cont=0)}	&48.6 /  {\bf 77.5}	&61.3 / 80.0	&36.5 / 77.3	&49.0 / {\bf 76.4}	&60.7 / {\bf 80.5}	&56.6 / {\bf 82.3}\\
\textbf{MMQL (cont=1)}	&51.0 / 77.4	&60.1 / {\bf 80.5}	&30.6 /{\bf  78.9}	&49.1 / 74.4	&58.0 / 79.6	&54.6 / {\bf 82.3}\\
\textbf{UNC }	        &48.4 / 69.1	&63.5 / 74.1	&19.4 / 53.5	&50.2 / 69.7	&59.0 / 71.9	&46.1 / 72.9\\
\textbf{RND }       	&48.2 / 70.4	&62.6 / 73.4	&21.5 / 56.7	&45.2 / 70.8	&59.4 / 73.9	&50.4 / 75.2\\
\hline \hline
\end{tabular}
\label{tab_views_1}
\end{table*}

\begin{table*}
\small
\caption{F-1 [\%] of the models before ({\it left}) and after ({\it right}) adaptation to test children using different budgets (5, 10, 20, 50 and 100) for active data-selection (averaged across the 14 children and 5 budgets).}
\begin{tabular}{l|c|c|c|c|c|c|}
\hline
\hline
{\textbf{Model}} &\textbf{FACE} &\textbf{BODY} &\textbf{A-PHYS} &\textbf{AUDIO}  &\textbf{FEATURE-F} &\textbf{MODEL-F} \\ \hline
\textbf{MMQL (cont=0)}	&28.9 / 42.2	&33.7 / 51.8	&23.2 / 44.1	&24.2 / 36.2	&33.2 / {\bf 48.5}	&32.0 / {\bf 52.6}\\
\textbf{MMQL (cont=1)}	&30.8 / {\bf 44.0}	&33.5 / {\bf 52.5}	&19.8 / {\bf 45.2}	&23.5 / {\bf 39.5}	&32.9 / {\bf 48.5}	&28.4 / 48.3\\
\textbf{UNC }	        &30.0 / 36.3	&35.6 / 41.9	&14.2 / 30.1	&24.3 / 32.0	&32.7 / 37.7	&28.1 / 38.1\\
\textbf{RND }       	&29.9 / 37.0	&35.0 / 41.6	&16.0 / 31.1	&24.2 / 32.7	&33.4 / 38.9	&29.6 / 38.9\\
\hline \hline
\end{tabular}
\label{tab_views_2}
\end{table*}

\subsection{Personalized Estimation of Engagement}
\label{per}
The learned group-policy for active data selection can be applied to multi-modal recordings of previously unseen children. However, the trained multi-modal engagement classifiers may be suboptimal due to the highly diverse styles of engagement expressions across the children with autism, in terms of their facial expressions, head movements, body gestures and positions, among others. These may vary from child to child not only in their appearance but also dynamics during the engagement episodes. To account for these individual differences, we adapt the proposed MMQL to each child. We do so by additionally training (fine-tuning) the modality-specific engagement classifiers using the actively selected data samples from the target child. Specifically, we start with the group-level engagement classifiers to obtain the initial engagement estimates, but also use the learned Q-function to select difficult data that need be expert-labeled. This is performed in an off-line manner: the requested samples of the target child are first annotated by an expert, and then used to personalize the engagement classifier. The main premise here is that with a small number of human-labeled videos, the engagement classifiers can easily be optimized for the target child as:
\begin{equation}
\varphi _*^{new} = \max \,{\langle \varphi _*^{(1)},\varphi _*^{(2)},..,\varphi _*^{(M)}\rangle _{D_{}^\ell }},
\end{equation}
where $D^{\ell}$ are the child data-samples actively selected using the group-level Q-function, and under the budget $\mathcal{B}$. This is described in detail in Alg.~\ref{pmmql}.

\section{Experiments}
\label{sec:exp}
{\bf Data and Features.}
To evaluate the proposed approach, we used the cross-cultural dataset of children with ASC attending a single session (on average, 25 mins long) of a robot-assisted autism therapy~\cite{rudovic2017measuring}. During the therapy, an experienced educational therapist worked on teaching the children socio-emotional skills, focusing on recognition and imitation of behavioral expressions as shown by neurotypical population. To this end, the NAO robot was used to demonstrate examples of these expressions. The data comprises highly synchronized audio-visual and autonomic physiological recordings of 17/18 children, ages 3-13, with Japanese/European cultural background, respectively. All the children have a prior medical diagnosis of ASC, varying in its severity. The audio-visual recordings were annotated by the human experts on a continuous scale [-1,1]. We discretized these annotations by binning the average continuous engagement score within 1 sec intervals into: low [-1,0.5], medium (0.5,0.8], and high (0.8,1] engagement. The multi-modal features were obtained as described in Sec.~\ref{deepf}. We split the children into training (20) and test (14)\footnote{The data of one child were discarded because of severe face occlusions.} at random, and used their recordings for the evaluation of the models presented here. Table~\ref{tab_views_3} shows the highly imbalanced nature of the data of the test children. 

\begin{figure}[b]
  \centering
 \includegraphics[width=0.5\textwidth]{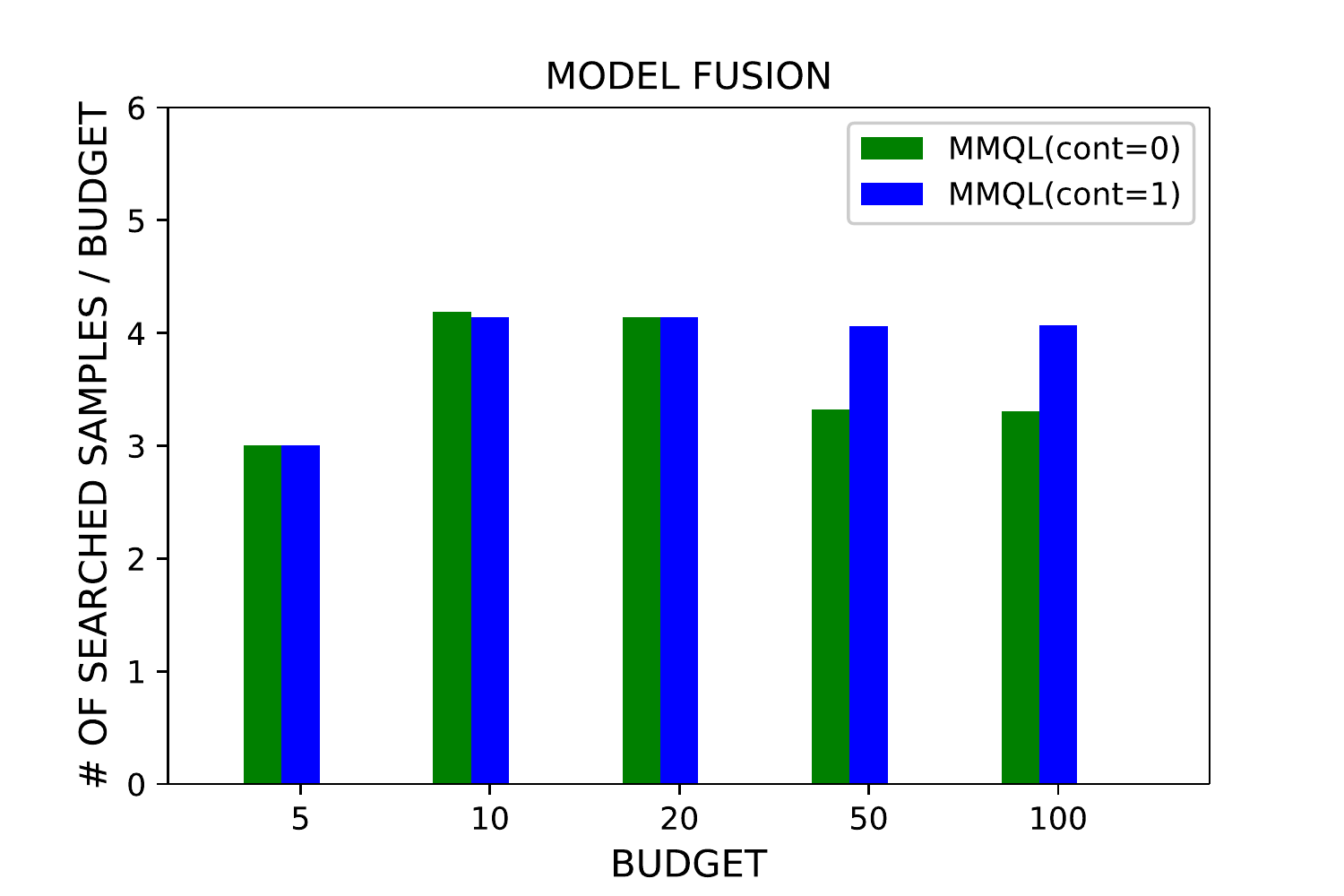}
  \caption{The relative number of searched data samples when the states of the Q-function in MMQL approach are: the raw input features {\bf x} (cont=1), and those constructed using the output of the modality-specific classifiers (cont=0).}
  \label{counts}
\end{figure}
\noindent{\bf Performance Metrics and Models.} We report the average accuracy (ACC) and F-1 score of the models in the task of 3-class engagement classification. The reported results are obtained by evaluating the group-level models and models adapted to each test child (Alg.~\ref{pmmql}). The latter was repeated 10 times by random shuffling of the test child data, and the average results are reported. For training/adaptation of the models, we varied the budget for active data selection as $\mathcal{B}\in\{5,10,20,50,100\}$. The number of episodes during training was set to 100, and LSTMs were trained for 10 epochs in each episode (and during the classifier adaptation to the test child). To evaluate different model settings, we start with the uni-modal models (thus, trained/tested using only $M=1$ data modality -- FACE, BODY, A-PHYS or AUDIO) and the multi-modal approach with the feature level fusion (i.e., by concatenating the input features from modalities $M=1,..,4$). We compare these models to the proposed model-level fusion in the MMQL approach. As the baselines, we show the performance achieved with alternative data-selection strategies: random sampling (RND) and the most common AL heuristic for data selection -- uncertainty sampling (UNC). In the case of multi-modal learning, the UNC scores for each sample were computed as the sum of the classifiers' entropy (Sec.\ref{gpolicy}).   
\begin{figure*}
  \begin{subfigure}[b]{0.4\textwidth}
  \centering
 \includegraphics[width=\columnwidth]{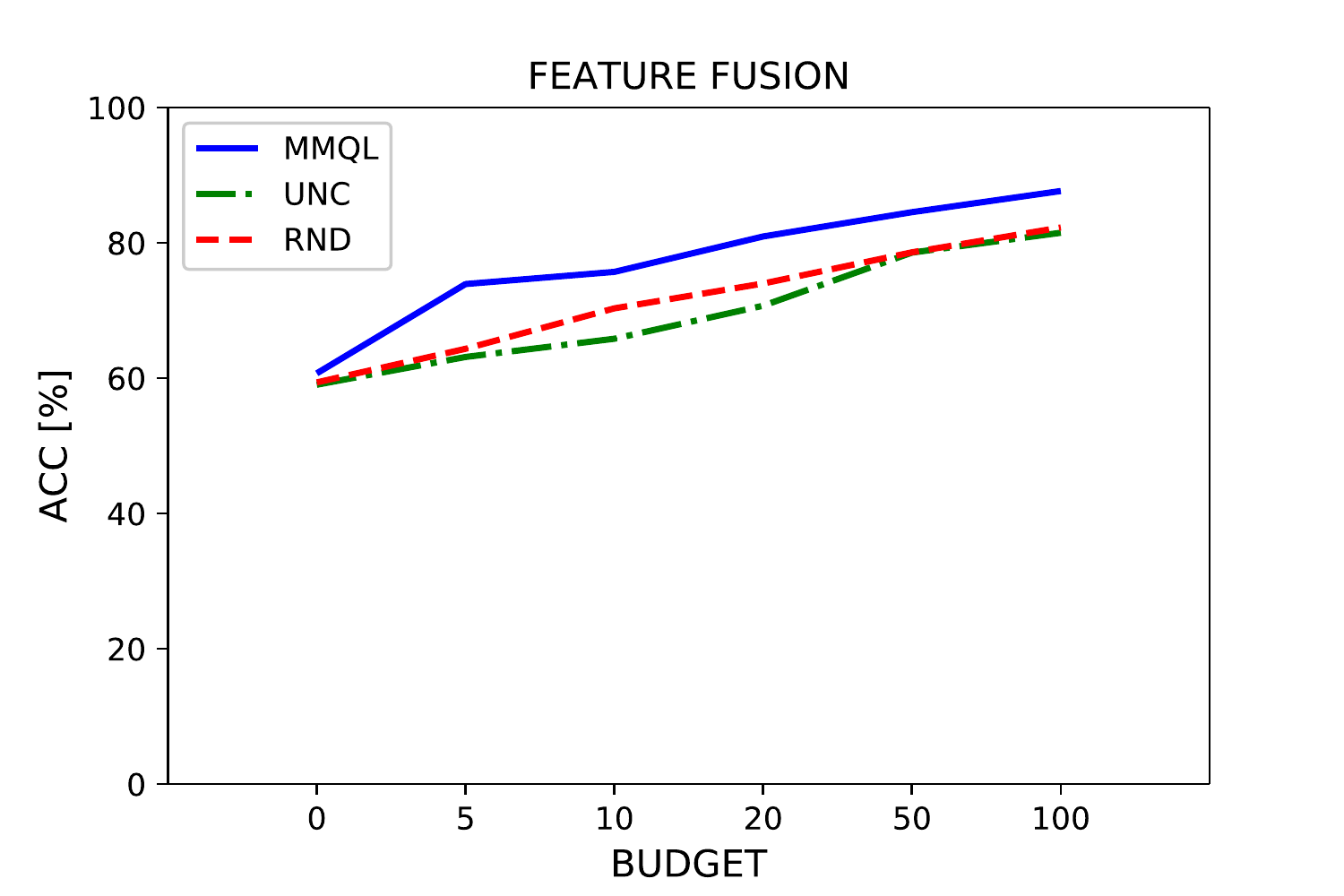}
  \end{subfigure}
\quad 
  \begin{subfigure}[b]{0.4\textwidth}
 \includegraphics[width=\columnwidth]{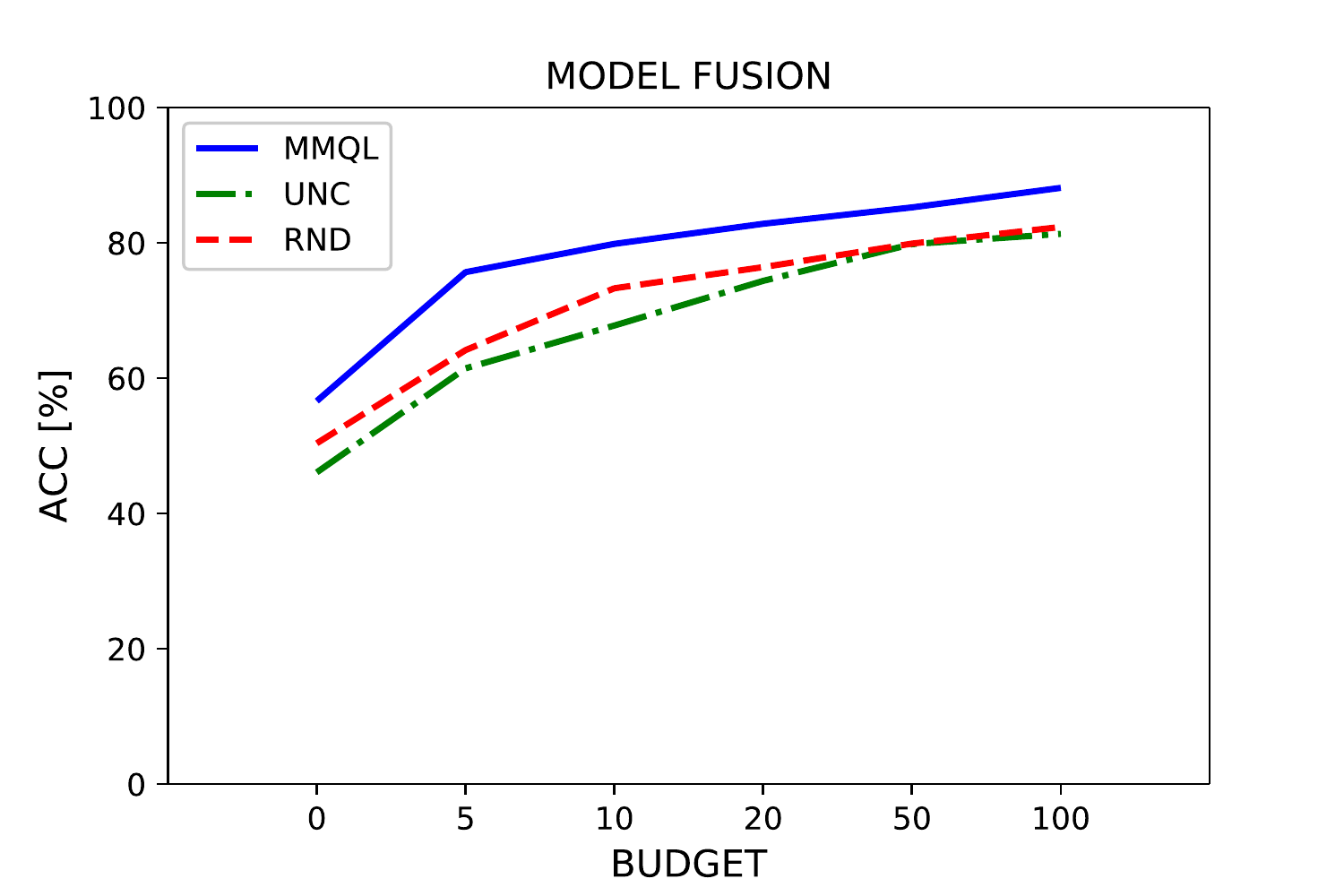}
  \end{subfigure}
 \\
   \begin{subfigure}[b]{0.4\textwidth}
  \centering
\includegraphics[width=\columnwidth]{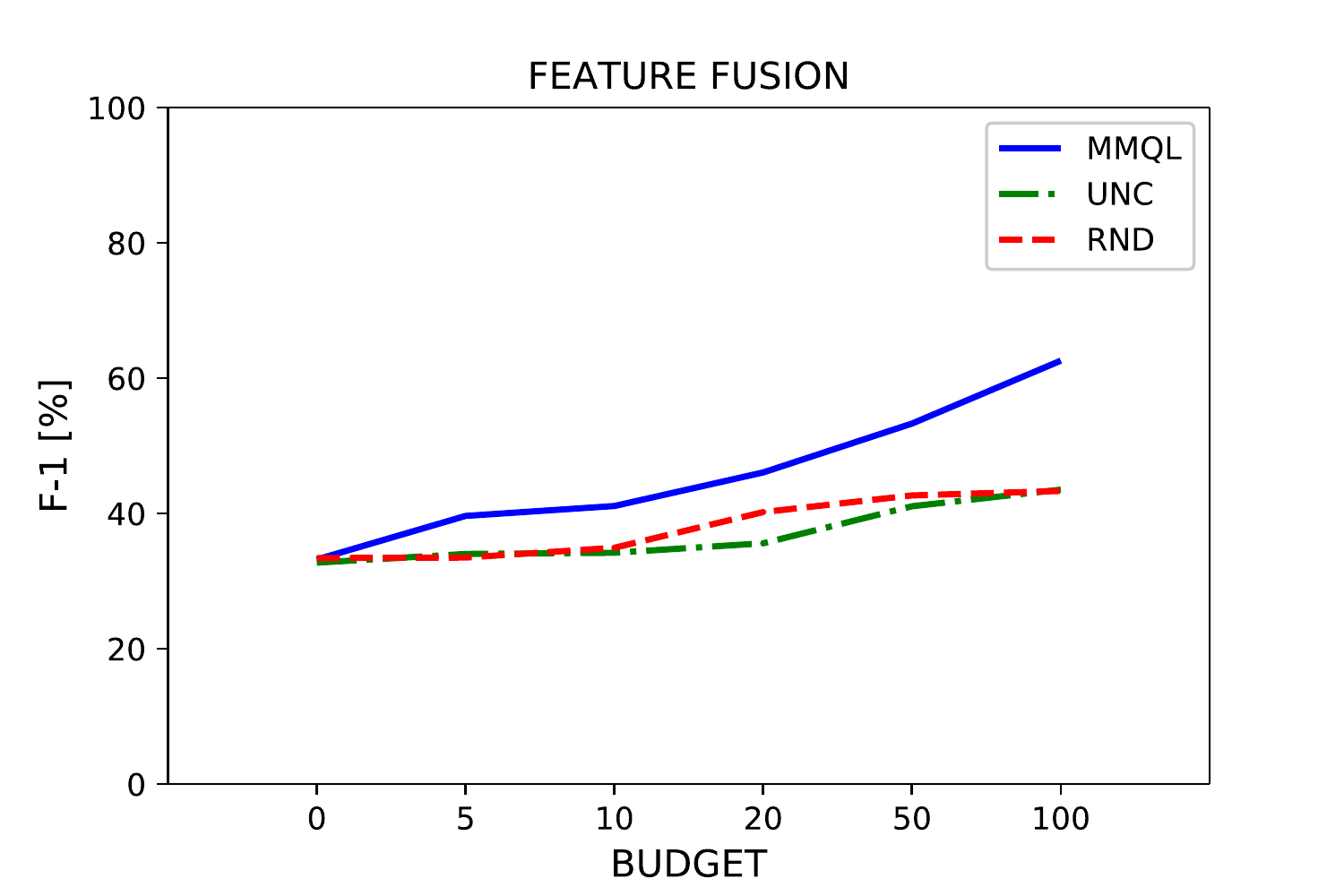}
  \end{subfigure}
\quad 
  \begin{subfigure}[b]{0.4\textwidth}
\includegraphics[width=\columnwidth]{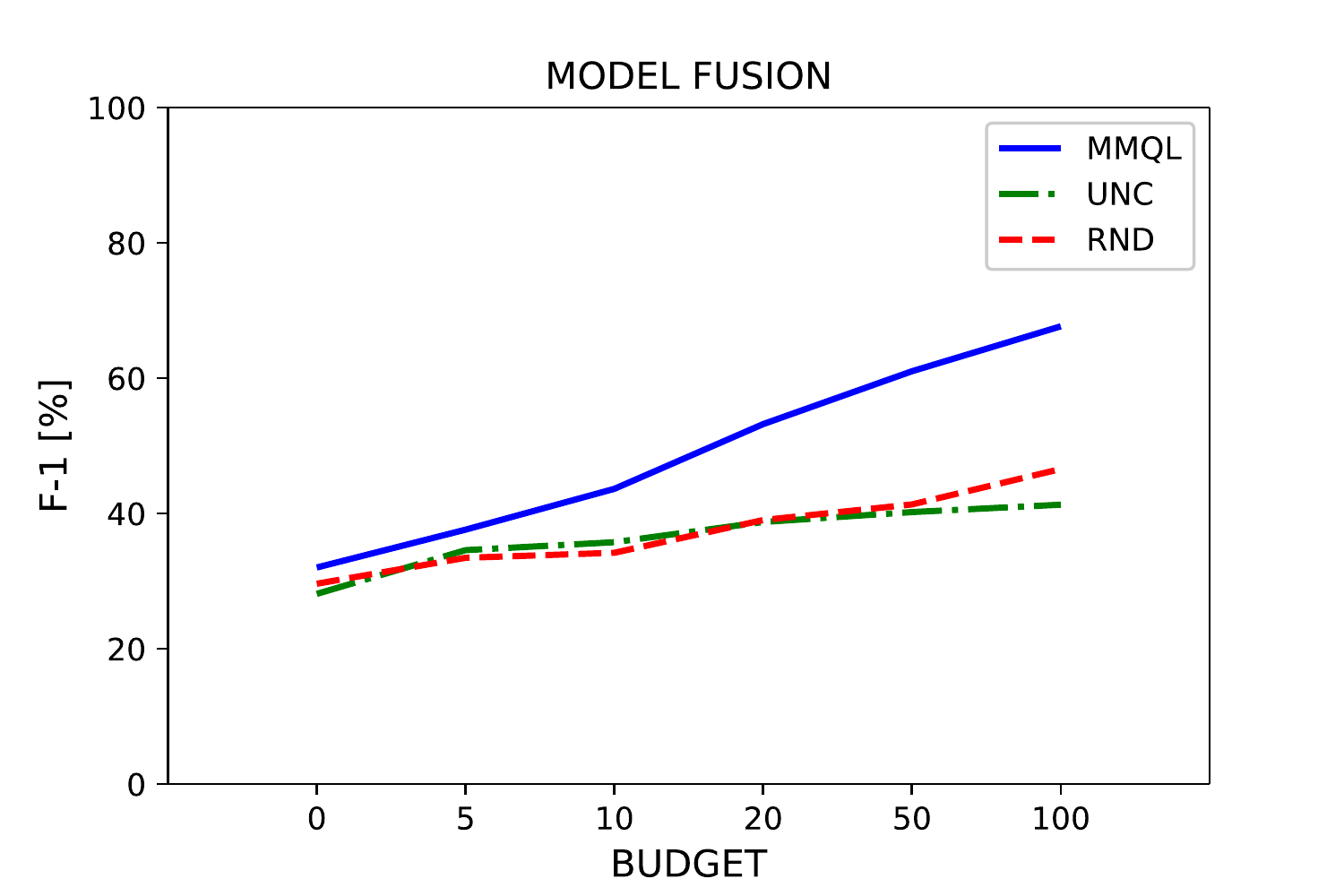}
  \end{subfigure}
  \caption{The performance of the feature- and model-level fusion with different active data-selection strategies.}
  \label{budgets}
\end{figure*}

\noindent{\bf Results.} Tables 2\&3 show the summary of the performances achieved by the compared models. We first note that the classifiers trained on the training children have generalized poorly on the test children. This is expected especially when using data of children with autism, who exhibit very different engagement patterns during their interactions. As can be seen from the numbers on the right-hand side (obtained after the personalization of the classifiers using actively selected data of the target child), the models' performances largely increase. This evidences the importance of the model personalizion using the data of the target child. Overall, among the uni-modal models, the BODY modality achieves the best performance, followed by the FACE, A-PHYS, and AUDIO modality, as also noted in~\cite{rudovic2018personalized}. However, both multi-modal versions of the models (feature- and model-level fusion) bring gains in the performance, with the model-level fusion performing the best on average. 

Comparing the MMQL models with the states based on the data content (cont=1) and those constructed from the classifiers outputs (cont=0), we note that there is no large difference in the performance for most models. Yet, the F-1 score of the MMQL with model-level fusion achieves a larger improvement (4.3\%). On the other hand, by looking at Fig.~\ref{counts}, we note that the MMQL approach with cont=0 requires a lower search time to reach the budget, while achieving a similar performance to when the content is used. Thus, this simpler model is preferable in practice. In the rest of the experiments, we show the performance of the MMQL (cont=0) only. Compared to the baselines, we note that the proposed MMQL largely outperforms these base strategies for active data-selection, under the same budget constraints. This evidences that the proposed is able to learn a more efficient data-selection policy. By comparing the ACC and F-1 scores, we note that the proposed is able to improve the classification of each engagement level, while the RND/UNC strategies tend to overfit the majority class.

\begin{figure*}
  \begin{subfigure}[b]{0.4\textwidth}
  \centering
\includegraphics[width=\columnwidth]{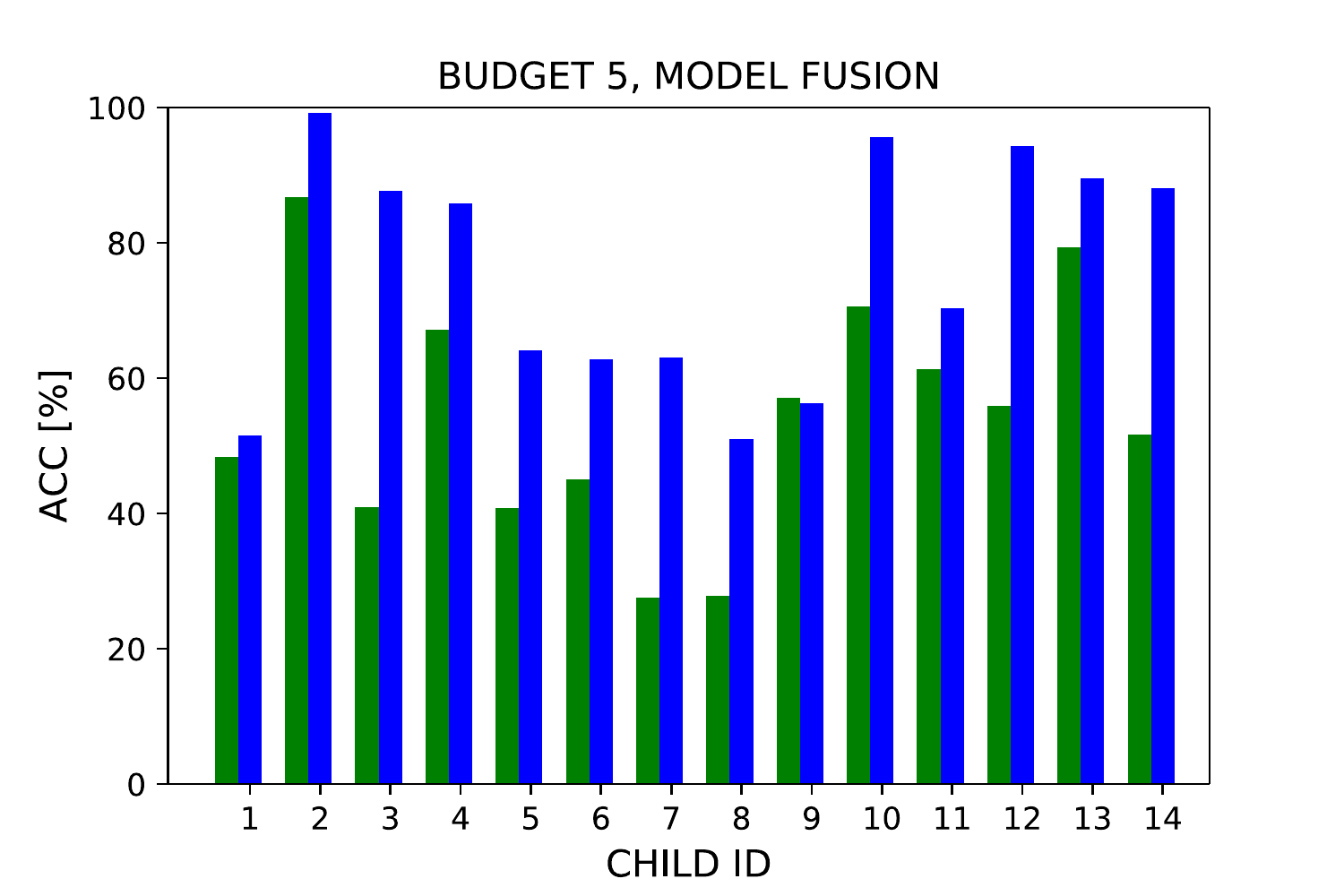}
  \end{subfigure}
\quad 
  \begin{subfigure}[b]{0.4\textwidth}
\includegraphics[width=\columnwidth]{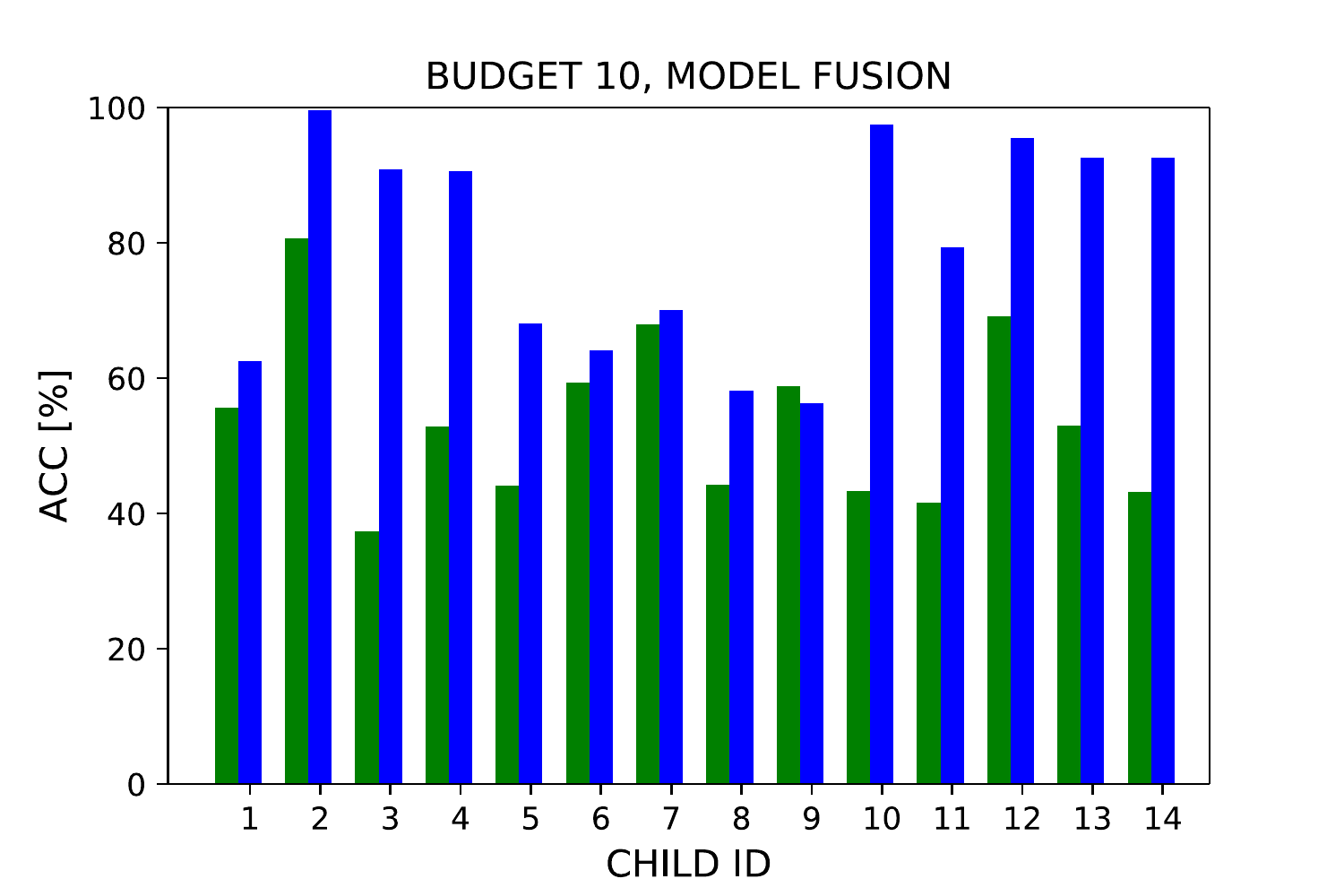}
  \end{subfigure}
 \\
   \begin{subfigure}[b]{0.4\textwidth}
  \centering
\includegraphics[width=\columnwidth]{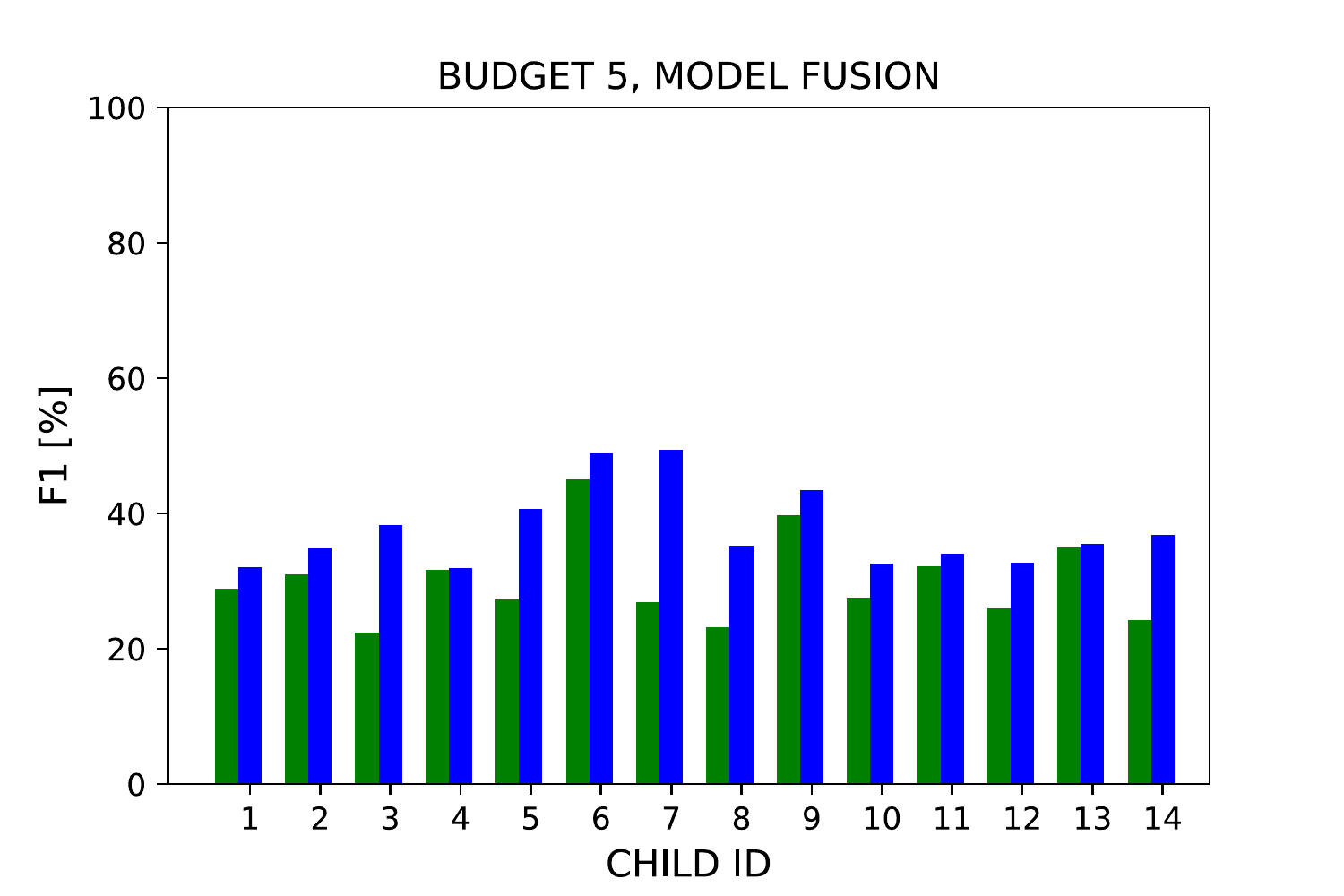}
  \end{subfigure}
\quad 
  \begin{subfigure}[b]{0.4\textwidth}
\includegraphics[width=\columnwidth]{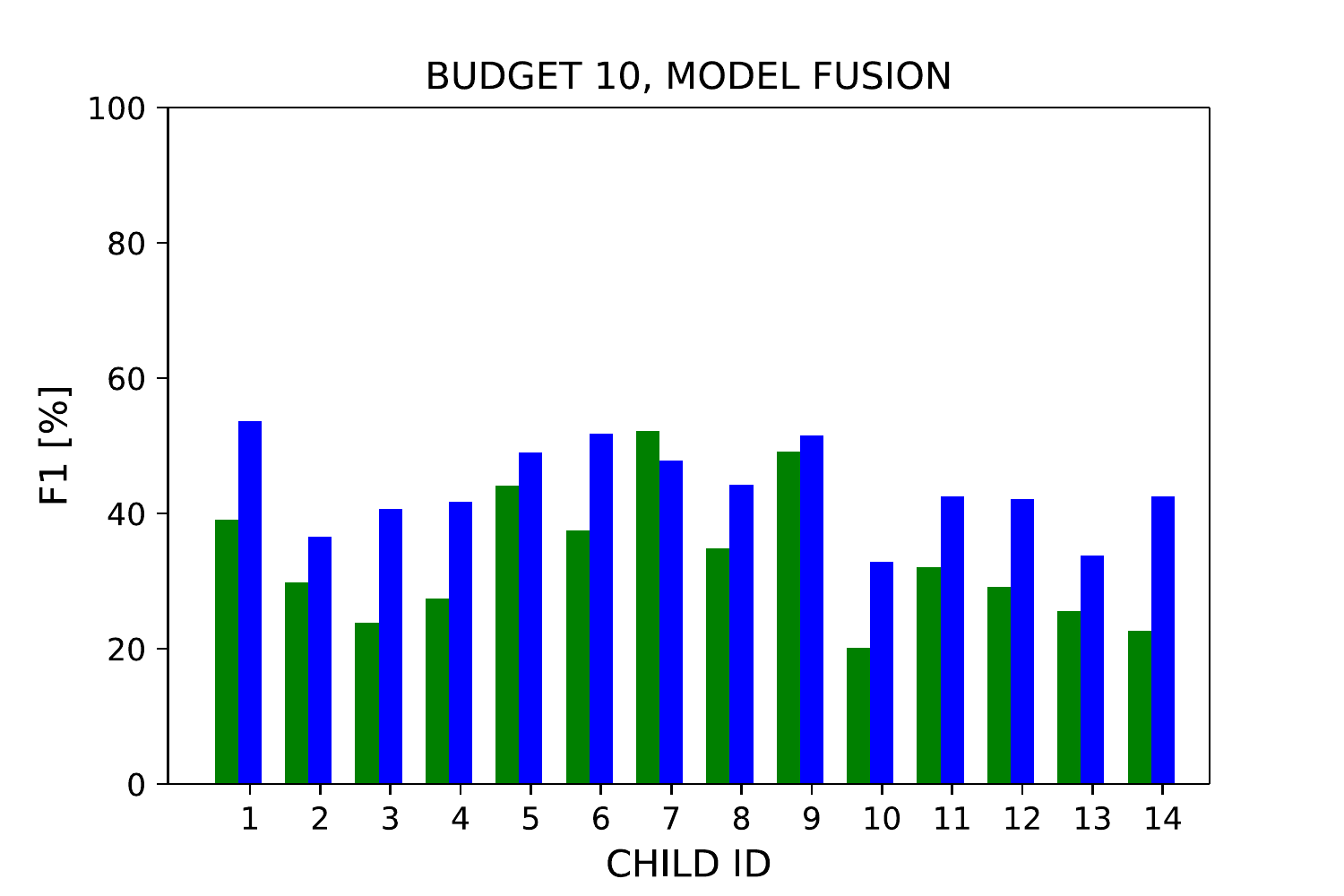}
  \end{subfigure}
\caption{The performance per test child of the MMQL (model-level fusion, cont=0) before (in {\it green}) and after (in $\it blue$) the personalization of the engagement classifiers. The results are shown for the budgets 5 and 10.}
\label{perchild}
\end{figure*}

Similar observations can be made from Fig.~\ref{budgets}, showing the performance of personalized models after the adaptation using different budgets. MMQL consistently outperforms RND and UNC sampling strategies, which we attribute again to the superior performance of the data selection strategy attained by the RL Q-function. This trend is even more pronounced for larger budgets, evidencing that the proposed Q-function consistently selects the more informative data samples, that are used to adapt the modality-specific classifiers to the target child. This holds for both, the feature- and model-level fusion within the proposed MMQL approach. 

Fig.~\ref{perchild} shows the performance of the MMQL (model-fusion, cont=0) approach per child before and after the model personalization using data actively selected with the proposed RL approach. Note that even with 5 samples only, the engagement classification improves largely and for almost all test children. However, because of the highly imbalanced nature of these data (see Table~\ref{tab_views_2}), F-1 scores are relatively low, yet, the improvements due to the active adaptation of the target classifiers are evident. One of the challenges when working with such imbalanced data is that the classifiers tend to overfit the majority class as most of the active samples come from that class. While this is much more pronounced in the RND/UNC selection strategies, it is also one of the bottlenecks of the current approach since the target classifiers are updated offline. We plan to address this in our future work. 
\section{Conclusions}
We proposed a novel active learning approach for multi-modal data of human behaviour. Instead of using heuristic strategies for active data-selection (such as the uncertainty sampling), our approach uses the notion of deep RL for active selection of the most informative data samples for the model adaptation to the target user. We showed the effectiveness of this approach on a highly challenging multi-modal dataset of child-robot interactions during an autism therapy. Specifically, we showed that the learned data-selection policy can generalize well to new children by being able to select their data samples that allow the pre-trained engagement classifiers to adapt quickly to the target child. We showed that this multi-modal model personalization can largely improve the performance of the engagement estimation for each test child using only a few expert-labeled data of the target child. 

{\small
   \bibliographystyle{abbrv}
  \bibliography{sample-base}
}
\end{document}